\documentclass[conference]{IEEEtran}
\IEEEoverridecommandlockouts
\usepackage{cite}
\usepackage{amsmath,amssymb,amsfonts}
\usepackage{graphicx}
\usepackage{textcomp}
\usepackage{xcolor}
\def\BibTeX{{\rm B\kern-.05em{\sc i\kern-.025em b}\kern-.08em
    T\kern-.1667em\lower.7ex\hbox{E}\kern-.125emX}}
\usepackage{fancyhdr}
\usepackage{xcolor,soul,framed}
\usepackage{xspace}
\usepackage{array}
\usepackage{eqparbox}
\usepackage{url}
\usepackage{longtable}
\usepackage{lipsum}
\usepackage{blindtext}
\usepackage{makecell}
\usepackage{mathtools}
\usepackage{commath}
\usepackage{multirow}
\usepackage{tabularx}
\usepackage[normalem]{ulem}
\usepackage{xspace}
\usepackage{algorithm}
\usepackage{algpseudocode}
\usepackage{amsfonts}
\usepackage{placeins}
\usepackage{amssymb}
\usepackage{multirow}
\usepackage{tabularx}
\usepackage{caption}
\usepackage{subcaption}
\usepackage{pifont}

\usepackage{xspace}
\usepackage{algorithm}
\usepackage{algpseudocode}
\usepackage{amsfonts}
\usepackage{placeins}
\usepackage{booktabs}
\usepackage[normalem]{ulem}
\definecolor{green}{RGB}{0, 150, 00}
\definecolor{orange}{RGB}{255, 120, 0}

\def\BibTeX{{\rm B\kern-.05em{\sc i\kern-.025em b}\kern-.08em
    T\kern-.1667em\lower.7ex\hbox{E}\kern-.125emX}}

\usepackage[pagebackref=true,breaklinks=true,colorlinks,bookmarks=false]{hyperref}
\hypersetup{
    colorlinks=true,
    filecolor=magenta,      
    urlcolor=magenta,
}

\usepackage{fancyhdr}
\hypersetup{
    colorlinks=true,
    filecolor=magenta,      
    urlcolor=magenta,
}

\begin{document}

\title{Active Learning with Contrastive Pre-training for Facial Expression Recognition}

\author{\IEEEauthorblockN{Shuvendu Roy, Ali Etemad}
\IEEEauthorblockA{Dept. ECE and Ingenuity Labs Research Institute \\
Queen's University, Kingston, Canada\\
\{shuvendu.roy, ali.etemad\}@queensu.ca}}

\maketitle
\thispagestyle{fancy}

\begin{abstract}
Deep learning has played a significant role in the success of facial expression recognition (FER), thanks to large models and vast amounts of labelled data. However, obtaining labelled data requires a tremendous amount of human effort, time, and financial resources. Even though some prior works have focused on reducing the need for large amounts of labelled data using different unsupervised methods, another promising approach called \textit{active learning} is barely explored in the context of FER. This approach involves selecting and labelling the most representative samples from an unlabelled set to make the best use of a limited `labelling budget'. In this paper, we implement and study 8 recent active learning methods on three public FER datasets, FER13, RAF-DB, and KDEF. Our findings show that existing active learning methods do not perform well in the context of FER, likely suffering from a phenomenon called `Cold Start', which occurs when the initial set of labelled samples is not well representative of the entire dataset. To address this issue, we propose contrastive self-supervised pre-training, which first learns the underlying representations based on the entire unlabelled dataset. We then follow this with the active learning methods and observe that our 2-step approach shows up to 9.2\% improvement over random sampling and up to 6.7\% improvement over the best existing active learning baseline without the pre-training. We will make the code for this study public upon publication at: \href{https://github.com/ShuvenduRoy/ActiveFER}{github.com/ShuvenduRoy/ActiveFER}.
\end{abstract}


\section{Introduction}
Facial expression recognition (FER) has seen growing interest in the deep learning community \cite{churamani2020clifer,kolahdouzi2021face,kolahdouzi2022facetoponet,sariyanidi2017learning,kollias2021affect} mainly due to its practical applications ranging from smart devices and medical care assistants to smart vehicles. However, the size of the labelled FER datasets is generally one of the concerns prohibiting further progress in the area. Many of the recently developed deep learning models, such as Transformer \cite{transformer}, inherently require very large amounts of data.
As a result, many recent works on FER have focused on developing methods that learn better representations from small amounts of labelled data \cite{roy2022analysis, CL_MEx, ST_CLR}. 

Although labelled datasets are hard to collect and annotate, unlabelled images are widely available on the Internet. Given a pre-defined labelling budget, the annotation process involves resolving the choice of which samples of the unlabelled dataset to annotate \cite{roy2022_ssl}. 
Recently, active learning has emerged as a viable solution for identifying key samples from the unlabelled set \cite{al_survey}. The basic idea of active learning is to start the training process with a few randomly selected samples and their corresponding labels. As the training progresses, a selection criterion is used to find more samples from the unlabelled set that are the best candidates for annotation. The model is then trained with this new labelled set along with the previously available labelled set. This cycle continues as long as the labelling budget is not exhausted.

A variety of new active learning methods have been proposed in recent years in different areas \cite{al_survey}. Although a few works have explored active learning specifically for FER \cite{yao2021action, ahmed2018wild}, more recent approaches in active learning \cite{Coresets, BADGE} have not yet been studied in this context. Moreover, to our knowledge, a comprehensive study to benchmark the performance of different active learning methods for FER under the same training protocol has not been conducted.

Another well-known fact about active learning training with a small labelling budget is the `\textit{cold start}' problem. The cold start problem occurs when the initial labelled set is either too small or not a good representative for the entire dataset. In such scenarios, the model fails to learn effective representations from the initial labelled set, and thus informative samples are not selected in later cycles of the active learning process. This may result in a final accuracy that is even worse than not using active sampling altogether. 
\begin{figure*}
    \includegraphics[width=0.99\textwidth]{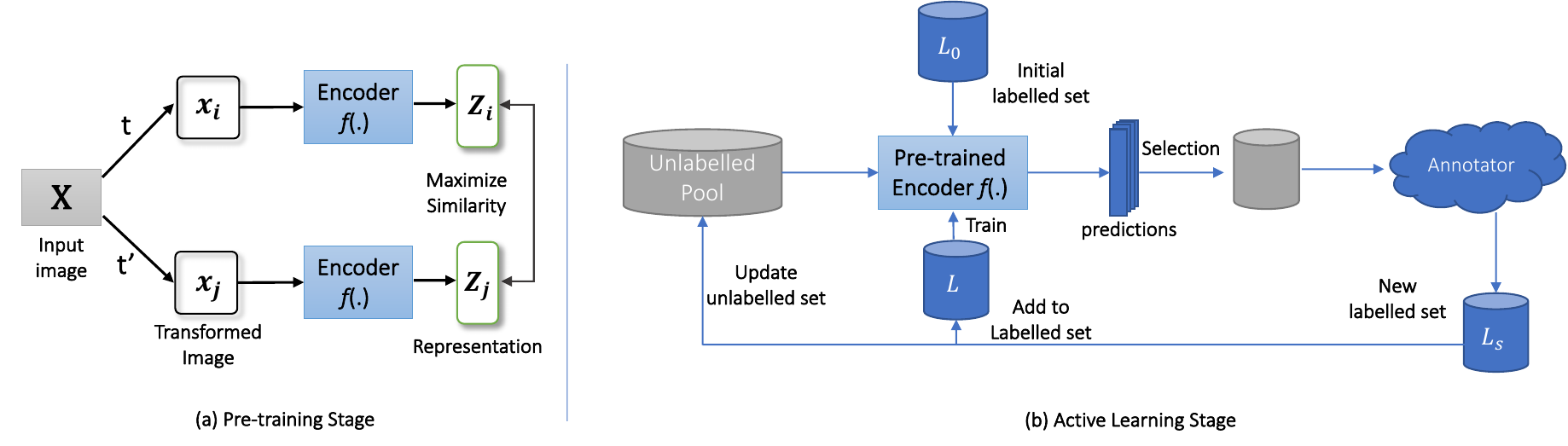}
    \caption{Overview of the 2-step training protocol. First, we pre-train the model with a contrastive self-supervised learning objective to learn the underlying representation of the data (left). Then we train it further with active learning (right). }
\end{figure*}

In this work, we address the two problems mentioned above. First, we present a comprehensive study of different active learning methods for FER. To this end, we compare 8 active learning methods, namely: Entropy \cite{entropy}, Margin \cite{margin}, Least Confidence \cite{margin}, BADGE \cite{BADGE}, GLISTER \cite{GLISTER}, Coreset \cite{Coresets}, BALD \cite{BALD}, and Adversarial Deepfool \cite{Deepfool}. We conduct our study on three FER datasets (FER13, RAF-DB, and KDEF) and show that surprisingly, simpler methods like Least Confidence and Margin obtain better results than recently proposed methods like Coresets and GLISTER. 
On average, the Least Confidence method shows the best performance across all three datasets. Additionally, we find that active learning on FER does indeed suffer from the cold start problem. To address this issue, we propose a simple yet effective solution: \textit{self-supervised pre-training} using the unlabelled data. We select the best-performing active learning method, Least Confidence, and show that by adding a self-supervised pre-training step, the cold start problem is reduced. Specifically, for self-supervised pre-training, we explore BYOL, MOCO, Barlow Twins, SwAV, and SimCLR and observe that while all of them are effective in reducing the negative impact of the cold start issue, SimCLR is the most effective.
The self-supervised pre-training step helps the method select more representative samples at the first cycle and effectively enables better learning at later cycles. 
Overall, our proposed solution shows up to 9\% improvement over random sampling and up to 6\% improvements over the scenario where the cold start issue is not addressed. 
Further ablation studies confirm that the improvements in performance are not simply due to a better encoder (pre-trained), but rather because of the fact that better samples are selected from the unlabelled set due to the added pre-training step. 

Our contributions in this work are summarized as follows:
\begin{itemize}
    \item We study active learning in the context of FER by exploring eight different active learning methods on three FER datasets. 
    \item We propose a new solution to reduce the cold start problem in active learning for FER, and show substantial overall improvements in performance.
    \item To contribute to the field of active learning in the context of affective computing and to enable reproducibility, we release the code for this work at: \href{https://github.com/ShuvenduRoy/ActiveFER}{github.com/ShuvenduRoy/ActiveFER}. 

\end{itemize}

\begin{figure*}
    \centering
     \begin{subfigure}[b]{0.49\textwidth}
         \centering
         \includegraphics[width=0.9\textwidth]{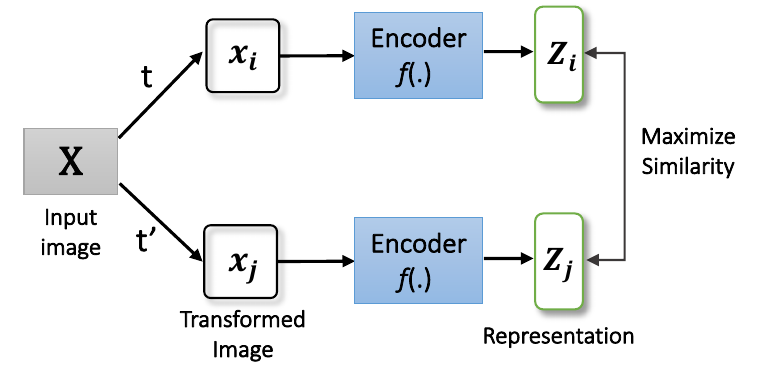}
         \caption{SimCLR}
         \label{fig:simclr}
     \end{subfigure}
     \hfill
      \begin{subfigure}[b]{0.49\textwidth}
         \centering
         \includegraphics[width=0.85\textwidth]{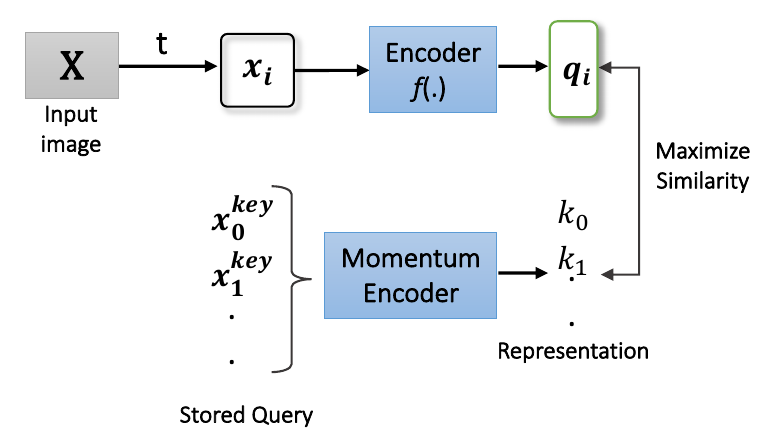}
         \caption{MoCO v2.}
         \label{fig:moco}
     \end{subfigure}
    \hfill
     \begin{subfigure}[b]{0.49\textwidth}
         \centering
         \includegraphics[width=0.9\textwidth]{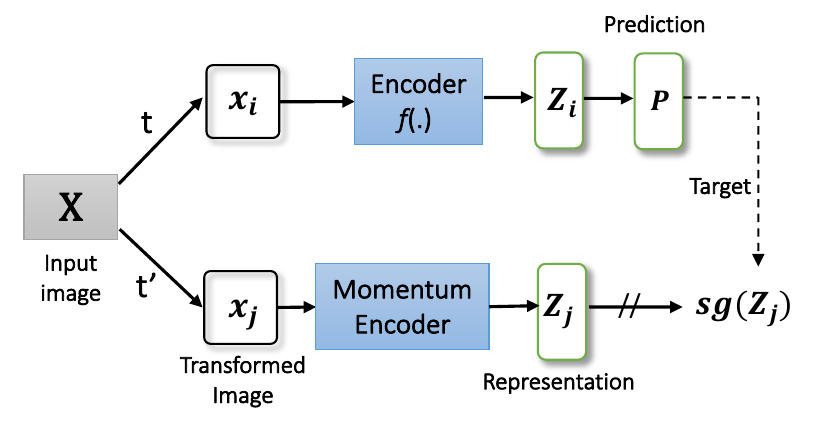}
         \caption{BYOL}
         \label{fig:byol}
     \end{subfigure}
      \hfill
     \begin{subfigure}[b]{0.49\textwidth}
         \centering
         \includegraphics[width=0.9\textwidth]{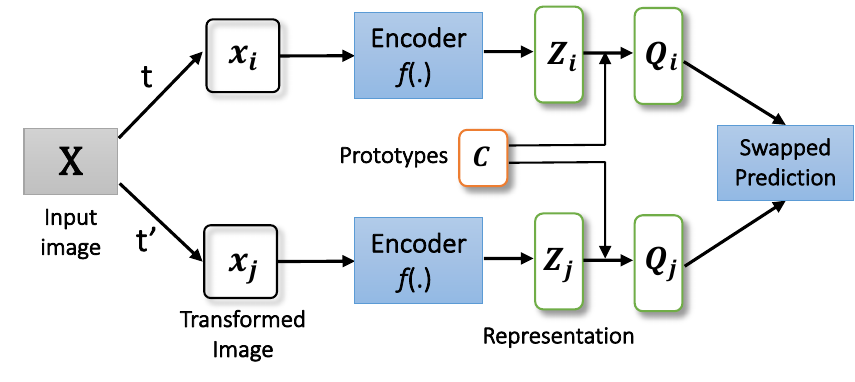}
         \caption{SwAV}
         \label{fig:swav}
     \end{subfigure}
      \hfill
     \begin{subfigure}[b]{0.49\textwidth}
         \centering
         \includegraphics[width=1.2\textwidth]{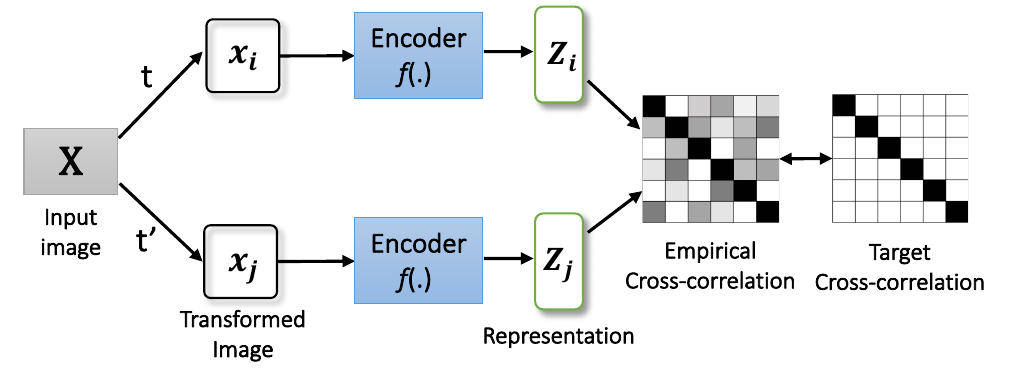}
         \caption{Barlow Twins}
         \label{fig:barlow}
     \end{subfigure}
    
    \caption{Architecture of different self-supervised training frameworks explored for the pre-training.}

    \label{fig:sensitivity}
\end{figure*}

\section{Related Work}

In this section, we discuss the related literature in two areas relevant to this work: (a) active learning and (b) self-supervised learning. 

\subsection{Active Learning}
The objective of active learning is to utilize a selection criterion for selecting the most representative samples from an unlabelled set for annotation. 
Although active learning is not a new concept, the rise of deep learning has resulted in a surge in active learning methods since deep learning methods require large datasets to train, which are not always available for many domains. 
Some selection methods in earlier forms of active learning utilized the concept of uncertainty in the model's prediction as an indicator for selecting new samples for labelling. For example, in \cite{entropy}, the entropy in the model's prediction on an unlabelled sample was used as the selection criterion. Two other variants of uncertainty-based sampling techniques were proposed in \cite{margin}. The first criterion (Margin) uses the difference between the top two predictions as the indicator for selection. A confident prediction will have a large difference between the highest and second-highest predictions over the number of classes. The second criterion (Least Confidence) simply takes the maximum over the class probability as the indicator. In this criterion, a less confident prediction will have lower confidence for the predicted class. 

More recent methods focus more on the learning progress of the model and utilize more specific signals as the selection criteria. For instance, \cite{BADGE} inspects the loss gradient and selects a set of samples with diverse loss gradients. GLISTER \cite{GLISTER} is another method focusing on diverse sampling over the entire dataset using bi-level optimization. The concept of Coresets \cite{Coresets} has also been utilized as a selection criterion. BALD \cite{BALD} uses Bayesian deep learning to maximize the information between the prediction and model posterior as an indicator for sampling. The concept of adversarial attacks has also been utilized to estimate the decision boundary of classes and select samples that are close to the boundaries \cite{Deepfool}.

\subsection{Self-supervised Learning}
Self-supervised learning (SSL) is one of the most popular unsupervised representation learning methods that has shown remarkable progress in various areas.
SSL can learn important representations of the data without any supervision. Most of the earlier SSL methods utilized the concept of pre-text tasks, where an auxiliary task was defined on the unlabelled data. One example of such a pre-text task is rotation prediction, where an input unlabelled image is rotated at a certain degree and the model is tasked with predicting the angle of rotation. A more recent form of SSL utilizes the concept of contrastive learning.
Contrastive learning in computer vision was popularized by SimCLR, which utilizes a contrastive loss on two augmentations of an unlabelled image to maximize their agreement in the embedding space. SimCLR also utilizes the concept of projection heads, hard augmentations, and carefully designed training protocols to perform effectively in many scenarios. 
Many variants of SimCLR have been since proposed. For instance, MoCo \cite{moco} utilizes a momentum encoder to encode one augmented image, while the other image is encoded by an online encoder. Later BYOL \cite{byol} proposed a slightly different objective function with respect to MoCO, which predicts the embedding of one augmented image from the other. SwAV \cite{swav} also has a similar idea with the distinction of predicting a prototype rather than the actual embedding. Barlow Twins \cite{barlow} proposed a different loss calculated on the cross-correlation between the predicted embedding of the two images. This method explicitly avoided the mode collapse problem of self-supervised learning while achieving strong performance on downstream tasks.
 
\section{Method}
\subsection{Preliminaries}
Let $X_U={(x_i)_{i=1}^N}$ be a set of unlabelled samples, where $N$ is the total number of samples in the unlabelled set, and $n$ be the total labelling budget, where $n \ll N$. An active learning method first randomly samples a small subset of $s$ samples from $X_U$ and annotates them to form $X_{L_S}={(x_i,y_i)_{i=1}^s}$. The model is trained with $X_{L_S}$ for certain epochs. These two steps are together called a cycle. Over the next $(c-1)$ cycles, the active learning method samples $(n-s)/(c-1)$ samples per cycle using a selection criterion, and trains the model with the current and previously sampled data. In this study, we explore the following active learning methods.

\textbf{Entropy.}
In Entropy, the uncertainty in the prediction of the model is used as the selection criterion \cite{entropy}. Let $p(x)=softmax(M_j(x))$ be the prediction of the model, where $M_j$ is the trained model at cycle $j$. The entropy selection criterion ($H(x)$) is represented as:
\begin{equation}
    H(x) = - \sum_i{p(x)_i log(p(x)_i)},
\end{equation}
where $i$ is the index over the vector dimension of the model's predictions. Here, the method chooses the sample with the highest entropy.

\textbf{Margin.}
This method considers the difference between the highest two predictions as the selection criterion ($F(x)$) \cite{margin}, which is defined as: 
\begin{equation}
    F(x)=p(x)_{m_1}-p(x)_{m_2}.
\end{equation}
Here, $m_1$ and $m_2$ are the largest and second-largest predictions, and the active learning method selects the sample with the lowest margin. 

\textbf{Least Confidence.}
Least Confidence is another simple approach where the prediction confidence is used as the selection criterion \cite{margin}. This criterion ($C(x)$) is defined as: 
\begin{equation}
    C(x)=max_i p(x)_i,
\end{equation}
Here, the active learning method selects the minimum $C(x)$ over all the unlabeled samples. 

\textbf{BADGE.}
This approach is one of the more recent active learning selection methods that take the model's learning progress into consideration \cite{BADGE}. It first computes the gradient of the last layer with some pre-defined loss function. Then, a K-Means++ algorithm is utilized to find the desired number of centers with diverse loss gradients.

\textbf{GLISTER.}
In this recent method, the active learning solution aims to select samples that are representative of the entire domain \cite{GLISTER}. For this, GLISTER utilizes a bi-level optimization problem where the inner optimization learns model parameters, and the outer optimization selects a set of unlabeled samples.

\textbf{Coreset.}
In this approach, the aim of the method is to find the samples that represent or capture the structure of the entire unlabelled dataset \cite{Coresets}. Here, a k-center algorithm is utilized to solve a pre-defined objective function to find the coresets.

\textbf{BALD.} In this method, the concept of Bayesian deep learning is utilized for active learning \cite{BALD}. BALD uses the concept of information maximization between the prediction and model posterior to select a pool of samples. 

\textbf{Adversarial DeepFool.} This active learning method operates by selecting points that are closer to the decision boundaries \cite{Deepfool}. Since calculating the actual decision boundary in the embedding space is difficult, this method proposes to use the concept of adversarial attacks to estimate it. For each sample, the number of adversarial perturbations required to flip the prediction is considered as the indication of how close it is to the decision boundary.

 \subsection{Proposed Solution for the Cold Start Problem}\label{sec:proposed}

In this section, we present a simple solution to address the cold start problem in active learning for improved performance in the context of FER. The proposed solution involves a two-step training protocol. In the first step, we pre-train the model with the entire unlabeled set $X_U$ in a self-supervised setting to learn the underlying representation of the data. In the second step, we follow conventional active learning training, where the learned representation from the first step helps the model to learn a discriminative representation at the first cycle from the small labelled subset $X_{LS}$, without overfitting.
This approach enables a better selection of representative samples in later cycles of the active learning training process, effectively reducing the cold start problem. We explore some recently proposed self-supervised methods, including SimCLR \cite{simclr}, MoCo v2 \cite{moco}, BYOL \cite{byol}, SwAV \cite{swav}, and Barlow Twins \cite{barlow}, for the first step.
For the second step, we use the Least Confidence as the active learning component. Below, we provide a brief overview of each of the self-supervised methods that we explore to evaluate our proposed solution.

\textbf{SimCLR} \cite{simclr} is a popular contrastive self-supervised technique responsible for popularizing contrastive learning in the field of computer vision. The basic idea behind this method is to learn from positive and negative samples, where positive samples are variations or transformations of an input sample, typically generated through augmentations. All other samples are considered negative with respect to the input sample. By bringing together positive samples and moving them away from negatives in the embedding space, contrastive learning allows for effective learning of the underlying representation of the data. The contrastive loss function for two positive samples, denoted as $i$ and $j$, is defined by:
\begin{align}
\mathcal{L}_{i, j} = -log\frac{exp(cos(z_i, z_j)/\tau)}{\sum_{k=1}^{2N} \mathbf{1}_{[k \neq i]} exp(cos(z_i, z_k)/\tau)},
\end{align}
where, $z_i$ is the embedding of the encoder,  $cos()$ is the cosine similarity function, and $\tau$ is a temperature parameter. A visual illustration of the SimCLR method is depicted in Figure \ref{fig:simclr}.

\textbf{MoCo} \cite{moco} is another popular self-supervised learning technique. Similar to SimCLR, the basic idea behind this technique is to learn representations that can differentiate between positive and negative samples. The positive pairs are again generated by data augmentation, while negative pairs are taken from a queue of samples that were stored at previous iterations of training. MoCo maintains two encoders, one `online' and the other `momentum'. The online encoder is updated after processing each minibatch, while the momentum encoder is updated using a moving average of the online encoder parameters. This ensures that the momentum encoder is always slightly behind the online encoder, enabling it to capture information from a larger amount of samples. The momentum encoder is updated with the following equation: 
\begin{align}
    \theta_k = m*\theta_k + (1-m)*\theta_q,
\end{align}
where $\theta_q$ and $\theta_k$ are the parameters of the online and momentum encoder and $m$ is the momentum coefficient. 
MoCo v2 is a simple extension of MoCo that utilizes the projection head and hard augmentation concepts introduced in SimCLR. Figure \ref{fig:moco} depicts the MoCo framework.

\textbf{BYOL} \cite{byol} also utilizes two encoders called online and target encoder. Like MoCo, the target encoder is a moving average of the online encoder. The objective of BYOL is to train an online encoder to predict the target encoder's representation of the same image under different augmentations. The online encoder is updated with the following loss function:
\begin{align}
\mathcal{L}_{i, j} = 2 - 2 \frac{<P, Z_j>}{||P||_2 \cdot ||Z_j||_2},
\end{align}
where $Z_j$ is the embedding generated by the target encoder, and $P$ is the prediction generated from the online encoder's representation. BYOL framework is depicted in Figure \ref{fig:byol}

\textbf{SwAV} \cite{swav} is another popular self-supervised learning technique, which is a clustering-based approach. It first generates multiple views on a sample image by applying augmentations and then predicts its cluster assignment. Considering each view as a query and other views as keys, SwAV \cite{swav} utilizes contrastive learning on its cluster assignment. This clustering-based approach is able to learn high-quality representations even when the dataset is highly diverse or has many classes. The SwAV method is visualized in Figure \ref{fig:swav}.

\textbf{Barlow Twins} \cite{barlow} proposed a loss function that explicitly avoids the collapse in self-supervised representation learning. It does so by calculating the cross-correlation matrix between two augmented images and making this matrix close to identity. The loss function of Barlow Twins is represented as follows:
\begin{align}
    \mathcal{L}_{BT} = \sum_i{(1-C_{ii})}^2 + \lambda \sum_i \sum_{j\neq i}{C_{ij}}^2,
\end{align}
where $C_{ij}$ is the cross-correlation between $i$th and $j$th images in a batch. Here, the first term is called the invariance term, and the second term is called the redundancy reduction term. The Barlow Twins method is shown in Figure \ref{fig:barlow}.

\section{Experiments and Results}
In this section, we describe the experimental setup and the results. First, we present the implementation details and dataset description. Then we discuss the results of different active learning methods for FER. Finally, we present the results of our proposed method and a details study of different aspects of the method.

\subsection{Datasets and Implementation Details}

The experiments in the paper are conducted on 3 popular expression recognition datasets: \textbf{FER13}, \textbf{RAF-DB}, and \textbf{KDEF}. These datasets were selected to cover various aspects of FER datasets, including dataset size (from small to very large), spatial resolutions (from low to high), and sources (lab condition vs. in-the-wild). FER13 \cite{fer13} is an in-the-wild dataset that was collected by the Google search API. All images in this dataset have an input resolution of 48$\times$48, and it contains seven expression classes with 28K and 7K images in the training and validation splits, respectively. RAF-DB \cite{raf_db} is another in-the-wild dataset containing 12K and 3K images for training and validation, respectively. The images in this dataset are re-scaled to a resolution of 96$\times$96. KDEF \cite{kdef} is a smaller dataset that was collected in a lab environment with comparatively higher-resolution images.

To ensure a fair comparison, all the active learning methods in the experiment were trained under the same training settings. Specifically, a ResNet-18 model is trained for seven cycles ($c$) with 40\% of the total labelled samples of the original dataset. The models were trained with an SGD optimizer with a learning rate of 0.01 and a batch size of 20. The pre-training was done for 400 epochs following the implementation details of SimCLR \cite{simclr}. The code is implemented in PyTorch and trained with Nvidia V100 GPU.

{\renewcommand{\arraystretch}{1.0}
    \begin{table}[t]
    \small
    \setlength    \tabcolsep{5pt}
    \caption{Accuracy of FER13, RAF-DB, and KDEF for different active learning methods averaged over three runs. All experiments are done on seven cycles.}
    \begin{center}
    \begin{tabular}{l|ccc }
    
    \hline
    \textbf{Methods} & \textbf{FER13}  & \textbf{RAF-DB}  & \textbf{KDEF} \\
    \hline
    Random Sampling	&      63.15$\pm$0.33	&      73.74$\pm$1.52	&      83.57$\pm$2.85	\\ \hline
    Entropy	\cite{entropy}&      65.86$\pm$0.33	&      76.81$\pm$1.51	&      80.78$\pm$1.76	\\
    Margin	\cite{margin}&      \textbf{65.87$\pm$0.62}	&      77.64$\pm$0.93	&      83.30$\pm$0.92	\\
    Least Confidence \cite{margin}	&      65.63$\pm$0.28	&      76.63$\pm$0.95	&      \textbf{86.03$\pm$1.44}	\\
    BADGE \cite{BADGE}	&      64.91$\pm$0.21	&      77.29$\pm$0.59	&      82.00$\pm$5.11	\\
    GLISTER	\cite{GLISTER}&      65.13$\pm$0.2	&      77.02$\pm$0.76	&      83.71$\pm$1.67	\\
    CoreSet	\cite{Coresets}&      64.99$\pm$0.09	&      75.76$\pm$0.96	&      81.94$\pm$1.75	\\
    BALD \cite{BALD}	&      65.58$\pm$0.17	&      74.79$\pm$1.06	&      79.69$\pm$1.42	\\
    Adv. DeepFool \cite{Deepfool}	&      64.81$\pm$0.38	&      \textbf{78.56$\pm$0.48}	&      79.96$\pm$5.21	\\ \hline
    Average & \textbf{65.35} & \textbf{76.48} & \textbf{82.20} \\ 
    \hline
    \end{tabular}
    
    \label{tab:al_fer}
    \end{center}
    \end{table}
}

\subsection{Performance of Existing Active Learning Methods on FER}
In this section, we evaluate the performance of various active learning methods for FER on the three aforementioned datasets. The results of this study are presented in Table \ref{tab:al_fer}. We observe three important findings that summarize the results from this study as follows:

{\renewcommand{\arraystretch}{1.0}
    \begin{table*}[t]
    \small
    \setlength    
    \tabcolsep{10pt}
    \caption{Performance of different approaches averaged over three runs.}
    \begin{center}
    \begin{tabular}{l|ccc }
    
    \hline
    \textbf{Method} & \textbf{FER13}  & \textbf{RAF-DB}  & \textbf{KDEF} \\
    \hline
    Random Sampling	&      63.15$\pm$0.33	&      73.74$\pm$1.52	&      83.57$\pm$2.85	\\ 
    Least Confidence	&      65.63$\pm$0.28	&      76.63$\pm$0.95	&      86.03$\pm$1.44 \\
        Pre-training + Random Sampling & 64.66$\pm$0.22	  &  76.92$\pm$1.01 & 	     85.96$\pm$2.42	\\
    Pre-training + Least Confidence (our approach) &  \textbf{67.11$\pm$0.70}	&      \textbf{79.56$\pm$0.64}	&      \textbf{92.77$\pm$2.17} \\
    
    \hline
    \end{tabular}
    
    \label{tab:al_cl}
    \end{center}
    \end{table*}
}

\noindent \textbf{(1) Existing methods perform poorly on FER.} 
In general, active learning on FER does not provide a reasonable improvement over random sampling. Only Least confidence and GLISTER show improvements over random sampling for all datasets. 
The average across the active learning methods shows some improvement over random sampling for FER13 and RAF-DB datasets (2.30\% and 2.74\%) but gets reduced performance for KDEF. This signifies the fact that FER requires more specialized active learning methods to get a reasonable improvement over random sampling. 

\noindent \textbf{(2) Simpler methods outperform more complicated and specialized methods.}
For FER, simpler approaches such as Entropy, Margin, or Least Confidence show improvements over the latest methods like BADGE, GLISTER, and CoreSet. 
Among the two methods that show improvements for all datasets (Least Confidence, GLISTER), Least confidence shows 2.48\%, 2.89\%, and 2.46\% improvements on FER, RAF-DB, and KDEF, while GLISTER shows 1.98\%, 3.28\%, and 0.14\% improvements. Thus we can conclude that the Least Confidence method is the best default choice for FER datasets for active learning. 

\noindent  \textbf{(3) Active learning does not work well on small datasets.} 
In our experiment on the smallest dataset, KDEF, we find no improvement for most of the active learning methods. Apart from Least Confident and GLISTER, all other methods perform below random sampling. 
We argue that this poor performance is caused by the cold start problem. Since KFEF is a small dataset, training starts with a very small-sized labelled set, thus overfitting without learning generalizable representations. 
In the next section, we present the results of our proposed solution that alleviates the cold start problem and improves the accuracy of all datasets, including the very small KDEF dataset.

\begin{figure*}
\centering
 \begin{subfigure}[b]{0.29\textwidth}
     \centering
    \includegraphics[width=1.0\textwidth]{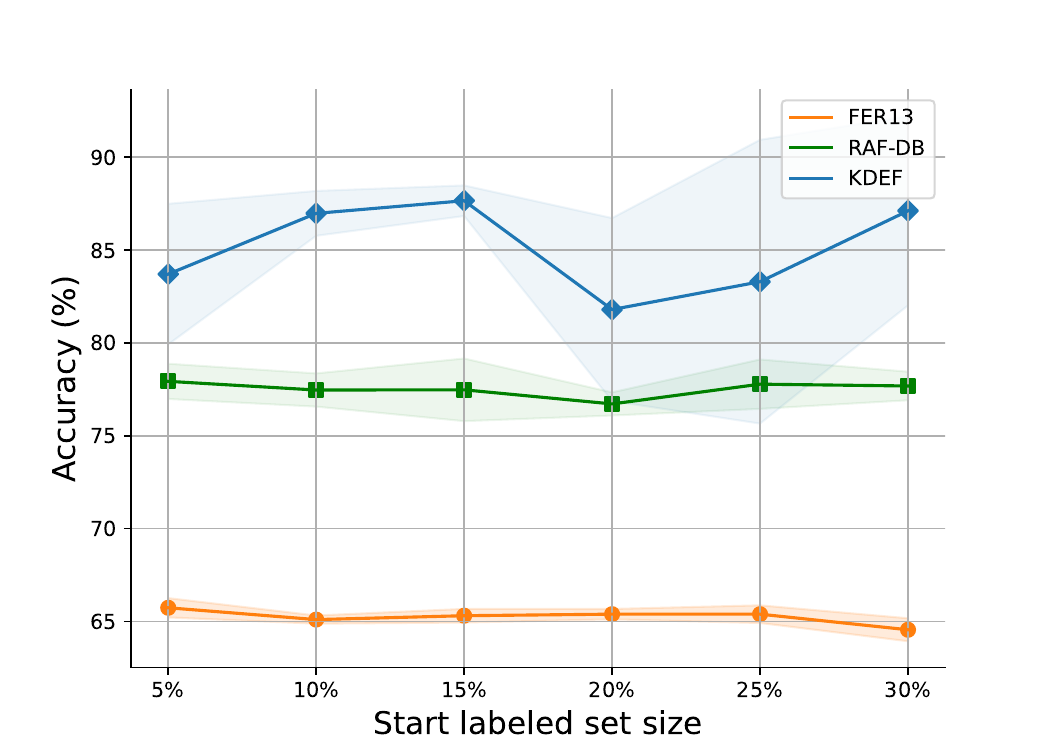}
     \caption{Initial labelled set size.}
     \label{fig:different_start_size}
\end{subfigure}
~~~~~~~~~~~~~~
\begin{subfigure}[b]{0.29\textwidth}
     \centering
     \includegraphics[width=1.0\textwidth]{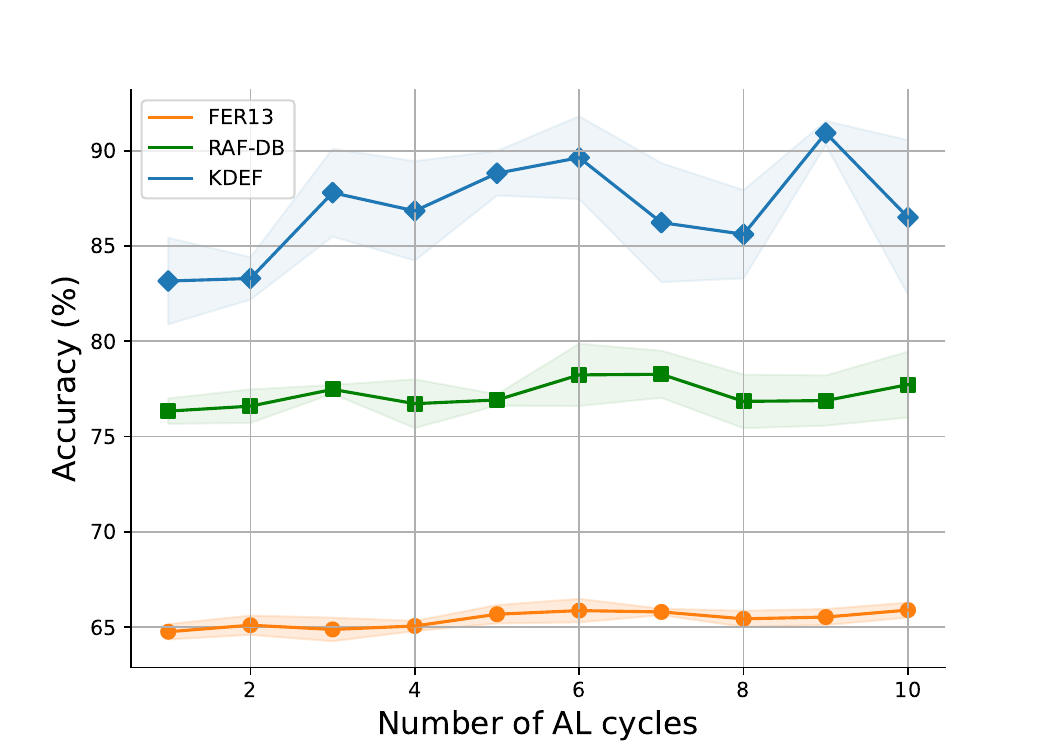}
     \caption{Number of cycles.}
     \label{fig:num_cycles}
 \end{subfigure}
 \\
 \begin{subfigure}[b]{0.29\textwidth}
     \centering
     \includegraphics[width=1.0\textwidth]{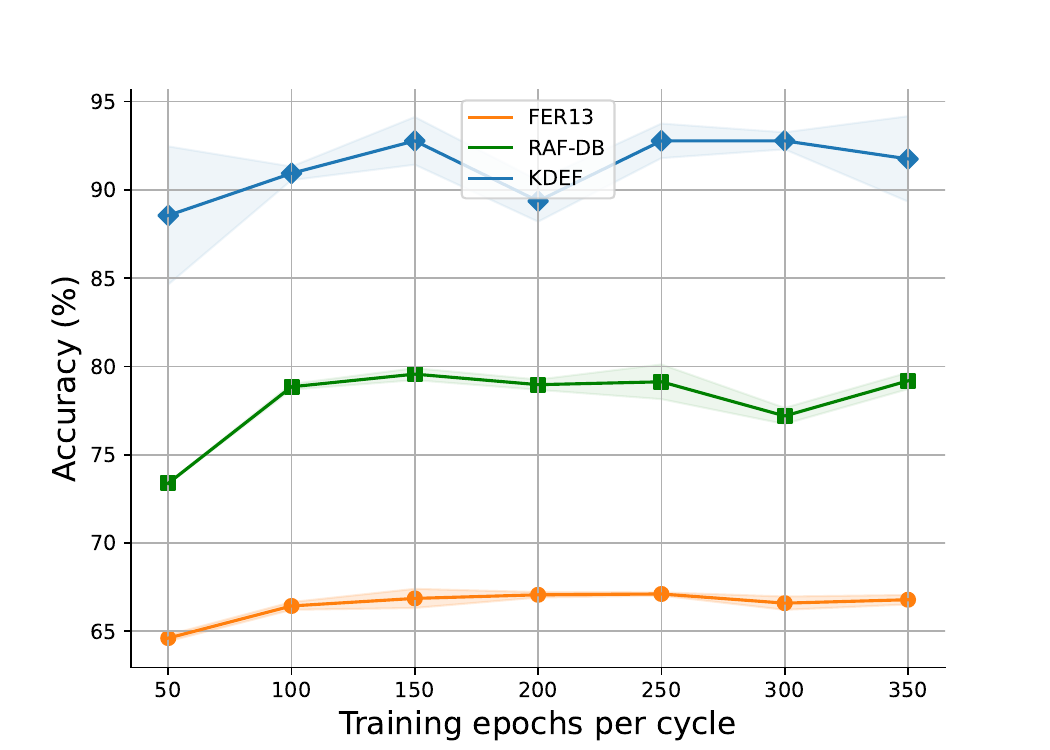}
     \caption{Epochs per cycle.}
     \label{fig:different_epochs}
 \end{subfigure} 
 ~~~~~~~~~~~~~~
 \begin{subfigure}[b]{0.29\textwidth}
     \centering
     \includegraphics[width=1.0\textwidth]{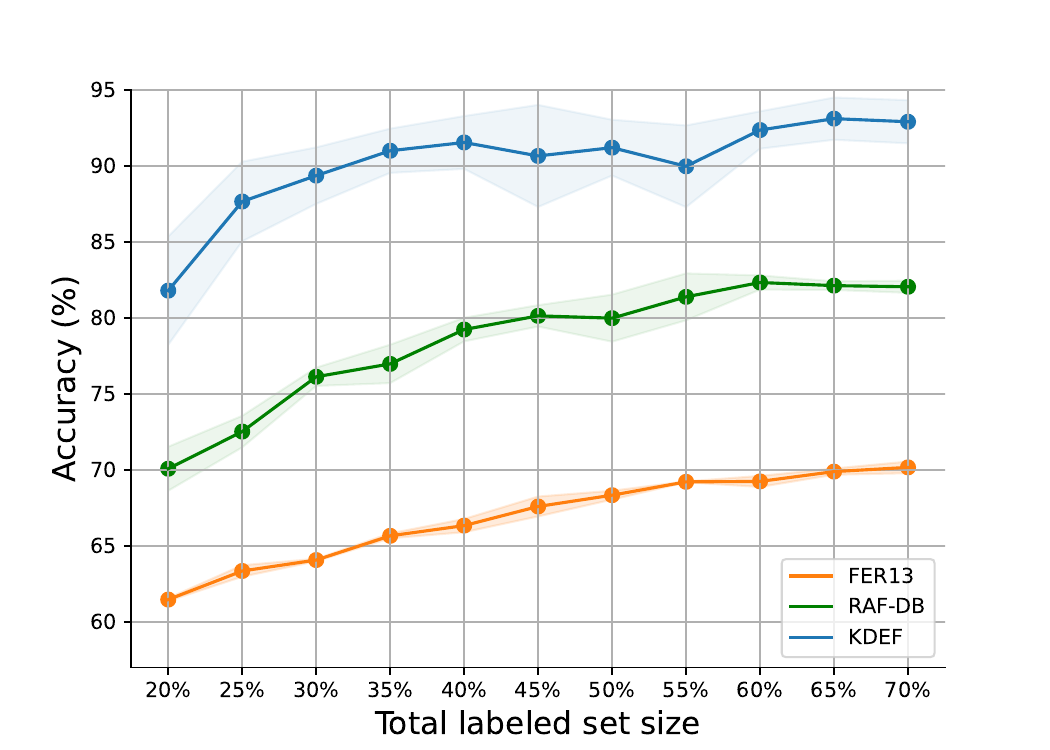}
     \caption{Total labelling budget.}
     \label{fig:different_total_samples}
 \end{subfigure} 
    \caption{Sensitivity study on various parameters of the 2-step training protocol on all three datasets. The light-shadowed area shows the standard deviation.}
    \label{fig:sensitivity}
    \vspace{-10pt}
\end{figure*}

\subsection{Performance of the Proposed Solution}
In Table \ref{tab:al_cl}, we present the performance of the best active learning method for FER, i.e., Least Confidence, with and without our proposed solution for self-supervised pre-training.
In general, we observe noticeable improvements with the proposed SSL pre-training in comparison to the baseline, with the highest improvement being 6.74\% on the KDEF dataset. Recall that most existing methods showed performance degradation on KDEF due to the cold start problem. Furthermore, the proposed method shows a 9.2\% improvement over random sampling on this dataset. This shows the effectiveness of the proposed approach for solving the cold start problem and improving performance. On the RAF-DB dataset, we see a 2.93\% improvement with pre-training and a 5.82\% improvement over random sampling. Similarly, for FER13, we observe a 1.48\% improvement with pre-training and 3.96\% over random sampling.

\subsection{Ablation Study}
In this section, we further analyze the impact of SSL pre-training on active learning. While Table \ref{tab:al_cl} demonstrated the positive impact of pre-training on the Least Confidence method, to investigate whether this improvement is a result of the pre-training alone or the fact that pre-training combined with active learning alleviates the cold start problem, we perform the same pre-training followed by random sampling. This result is presented in Table \ref{tab:al_cl}, where we observe a 2.45\%, 2.64\%, and 6.81\% drop in performance on the three datasets, respectively. This finding enforces that a pre-trained encoder (in a self-supervised setting) on its own does not provide much improvement. Rather, selecting more representative samples in active learning boosts performance.

We also investigate different choices for self-supervised pre-training, which we discussed in Section \ref{sec:proposed}. We summarize these results in Table \ref{tab:al_self}. Overall, observations from the table show that pre-training with SimCLR provides the best results compared to other self-supervised methods. Barlow Twins shows the next best accuracy for RAF-DB and KDEF. MoCo-v2 shows the second-best accuracy for FER13. While SimCLR provides 1.48\%, 2.93\%, and 6.74\% improvements for FER13, RAF-DB, and KDER datasets, Barlow Twins shows 2.73\% and 5.04\% improvements for RAF-DB and KDEF datasets, and MoCo v2 demonstrates 1.28\% improvements for FER13.

{\renewcommand{\arraystretch}{1.0}
    \begin{table}[t]
    \small
    \setlength    \tabcolsep{5pt}
    \caption{Performance of Least Confidence method with various pre-training approaches.}
    \begin{center}
    \begin{tabular}{l|lll }
    
    \hline
    \textbf{Pre-training} & \textbf{FER13}  & \textbf{RAF-DB}  & \textbf{KDEF} \\ \hline
    
    None	&      65.63$\pm$0.28	&      76.63$\pm$0.95	&      86.03$\pm$1.44 
    \\ 
    BYOL & 65.91$\pm$0.79 & 78.46$\pm$1.78 & 87.32$\pm$0.73\\
    MoCo v2 & 66.91$\pm$0.92 & 79.11$\pm$1.69& 90.80$\pm$1.04\\
    Barlow Twins & 66.85$\pm$0.87 &79.36$\pm$1.34 & 91.07$\pm$0.98\\ 
    SwAV  &66.02$\pm$0.99 & 78.12$\pm$1.95 & 88.89$\pm$1.79\\ 
    SimCLR & 67.11$\pm$0.70  & 79.56$\pm$1.64 & 92.77$\pm$2.17\\ 
    
    \hline
    \end{tabular}
    
    \label{tab:al_self}
    \end{center}
    \end{table}
}

\subsection{Sensitivity study} 
In this section, we present a detailed sensitivity study on different hyper-parameters involved in the entire pipeline, including pre-training and active learning. More specifically, we conduct experiments on the following factors: (1) initial labelled set size of active learning, (2) number of active learning cycles, (3) number of  epochs per cycle, (4) total labelling budget, and (5) optimizer. We show the results for the best performing active learning method for FER, i.e., Least Confidence, on all three datasets.

\subsubsection{Sensitivity toward the initial labelled set size} 
The `initial labelled set size ($s$)' is an important hyper-parameter for active learning. Since the initial labelled set is selected randomly, selecting a large value for $s$ reduces the number of samples that can be selected with active learning at later cycles. On the other hand, choosing very small values of $s$ can contribute to the cold start problem. Therefore, we investigate different values of $s$ for FER using the two-step training solution. In Figure \ref{fig:different_start_size}, we illustrate the sensitivity for different initial labelled set sizes. The figure shows that choosing small values of $s$ leads to better performance. For example, both FER13 and RAF-DB show the best performance when only 5\% of samples are selected as the initial labelled set. For KDEF, the best accuracy is achieved with an initial labelled set size of 15\%. We argue that the underlying representations learned by the self-supervised pre-training are responsible for this phenomenon. Due to the pre-training step, the model can learn from a small initial labelled set and select most of the samples in later cycles. Another important trend is the increased standard deviation when more samples are selected as the initial labelled set, especially for the KDEF dataset.

\subsubsection{Sensitivity toward the number of active learning cycles} 
Another important parameter for active learning training is the total number of cycles. A large number of cycles provides the opportunity for selecting more representative samples at later cycles. Nevertheless, increasing the number of cycles also increases the total training cost, and it is, therefore, important to find an optimal number for this parameter. The sensitivity toward different numbers of training cycles is presented in Figure \ref{fig:num_cycles}. We find the best performance is obtained for FER and RAF-DB when the model is trained for seven cycles and KDEF for nine cycles. 

\subsubsection{Sensitivity toward training epochs per cycle} 
We also investigate the optimal number of training epochs per cycle. 
This is another important parameter since training for an excessive number of epochs can cause the model to overfit the data available in that cycle, whereas using too few can hamper learning for the set of data in that cycle. As a result, we present a sensitivity study on the number of active learning epochs in Figure \ref{fig:different_epochs}. The observations show that the best accuracies are observed for 150 epochs for both KDEF and RAF-DB datasets. The FER13 shows better performance for longer training as it obtains the best accuracy with 250 epochs, with 200 also showing a close accuracy.

\subsubsection{Performance versus different labelling budgets} 
In Figure \ref{fig:different_total_samples}, we show the performance for different amounts of total labelling budget. In general, more labelled samples results in better accuracy for almost all settings. However, the change in accuracy is sharper in low data regimes (e.g. increase from 20\% to 25\%) compared to the higher ones.

{\renewcommand{\arraystretch}{1.0}
    \begin{table}[t]
    \small
    \setlength    \tabcolsep{5pt}
     
    \caption{Accuracy for different optimizers.}
    \begin{center}
    \begin{tabular}{l|ccc }
    
    \hline
    
    \textbf{Optimizer} & \textbf{FER13}  & \textbf{RAF-DB}  & \textbf{KDEF} \\
    \hline
    
    SGD &   67.11$\pm$0.70 & 79.56$\pm$1.64 & 92.77$\pm$2.17 \\
    Adam &      63.67$\pm$0.68	&      75.25$\pm$2.97	&      85.48$\pm$2.32	\\

    \hline
    \end{tabular}
    \label{tab:optim}
    \end{center}
    \vspace{-20pt}
    \end{table}
}

\subsubsection{Sensitivity toward the optimizer} 
We also investigate the impact of the choice of optimizer on the final performance in Table \ref{tab:optim}. The results in this table show that SGD is considerably better than Adam optimizer for all the datasets. SGD shows 7.29\%, 4.31\%, and 3.44\% higher accuracy compared to Adam on KDEF, RAF-DB and FER13, respectively.

\section{Conclusion}
In this study, we explored active learning as a solution for reducing the reliance on large amounts of labelled data to train deep learning models for FER.  First, we implemented and evaluated various active learning methods for FER and confirmed the presence of a cold start problem. To overcome this issue and further enhance FER performance, we employed a two-step training protocol. In the first step, we conducted contrastive self-supervised pre-training using the entire set of unlabeled data. Our extensive studies showed that the two-step training protocol alleviates the cold start problem and improves performance by considerable margins. An extensive ablation study showed the effectiveness of the proposed pre-training step, and a comprehensive sensitivity study identified the optimal parameters for each dataset. We hope that this research will bring attention to this important direction for reducing the labelling cost, which can help facilitate the development of improved FER methods.

\section*{Acknowledgements}
We would like to thank Bank of Montreal and Mitacs for funding this research. We are also thankful to SciNet HPC Consortium for helping with the computation resources.

 \section*{Ethical Impact Statement}

This study did not include the collection of any new datasets as it relied on three popular public FER datasets used by most existing works in this domain. These datasets feature diverse demographic groups and contain no personal identification information (besides what has been already public or known to the original authors of these datasets) or offensive material, thereby avoiding any privacy concerns. Since our work relies on datasets collected from the internet, we believe good generalizability towards different demographic groups of people is achievable.
However, a detailed study on bias is required to further analyze this notion.
We acknowledge that, like any FER method, the system developed in this paper has the potential to be used to analyze the facial expressions of individuals without their consent. As a result, we find it absolutely imperative for such systems to be used ethically and responsibly with full compliance with ethical, moral, and legal guidelines. 
The training process of each experiment took under 12 hours on a single Nvidia V100 GPU, a reasonable time-frame that does not pose a significant carbon footprint.

\bibliographystyle{IEEEbib}
\bibliography{ref.bib}

\end{document}